# Effects of GIMP Retinex Filtering Evaluated by the Image Entropy

A.C. Sparavigna[1] and R. Marazzato[2]
1 Department of Applied Science and Technology, Politecnico di Torino, Torino, Italy
2 Visiting Staff at Department of Mathematical Sciences, Politecnico di Torino, Torino, Italy

**Abstract:** A GIMP Retinex filtering can be used for enhancing images, with good results on foggy images, as recently discussed. Since this filter has some parameters that can be adjusted to optimize the output image, several approaches can be decided according to desired results. Here, as a criterion for optimizing the filtering parameters, we consider the maximization of the image entropy. We use, besides the Shannon entropy, also a generalized entropy.

In 1948, Claude Shannon defined the entropy H of a discrete random variable X, as the expected value of its information content, that is, $H(X) = \sum_i p(x_i) I(x_i) = -\sum_i p(x_i) \log_b p(x_i)$ [1,2]. In this expression, I is the information content of X, the probability of i-event is $p_i$ and b is the base of the used logarithm (common values are 2 or Euler's number e). Therefore, the entropy is a measure of the information contained in a sample set of possible events. If we consider an image and its grey tones, we can calculate an entropy by using the histogram of the tones, instead of the probability distribution. In this manner, we have the "image entropy". In the calculus of image entropy, the Shannon entropy can be used but also the generalized entropies of Tsallis and Kaniadakis [3-6] (in the Appendix, some formulas are given).

Since each image has its entropy, when we apply a filtering to the image, its histogram is modified, and then the entropy is changed. If the filtering has some adjustable parameters, we have to choose them according to a suitable best choice. In [7] for instance, we proposed for the GIMP Retinex, a filter of the GNU Image Manipulation Program, for enhancing foggy images; this filter requires a choice of parameters, and a consequent ranking of output images, that we have based on the local variance of grey tones. Here, we consider the use of the entropy, on the images obtained by a Retinex filter. In this approach, the best choice of parameters is that corresponding to a filtered output image, which is maximizing entropy, then maximizing information contained in the image.

The GIMP Retinex is Multi-Scale Retinex with Colour Restoration (MSRCR) [8], which is a freely available tool developed by Fabien Pelisson [9]. The output image of G-Retinex can be adjusted selecting different levels, scales and dynamics. We have three "levels" of filtering. The uniform level is treating both low and high intensity areas in the same manner. The low level enhances the lower intensity areas of the image whereas the high level is favouring the clearer areas. Another parameter of the filter is the "scale", which determines the depth of the Retinex scale. A "scale division" determines the number of iterations in the multiscale Retinex filter. The last parameter is a "dynamic" slider, which allows adjusting color saturation contamination around the new average color. The default values of scale, scale division and dynamic slider are 240, 3 and 1,2 respectively.

In this preliminary discussion, we use a few images just to test the possible use of entropy. We will calculate the Shannon entropy of the image and the Kaniadakis entropy, in the range of its entropic parameter [0,0.1] (see the appendix for formulas). Let us remember that the Shannon entropy is the limit value of Kaniadakis entropy when the entropic index approaches zero. Let us start from four images given in the Figure 1. We can see the original image and the three images obtained after filtering with GIMP Retinex setting the default values, in the case of the three levels, uniform, low and high. Looking at Figure 1 it seems that it is the low level the possible best choice. In fact, as we can see in the plot of Figure 2, the image obtained with the filter at low level has the highest value of entropy.

Of course, besides the choice of level, we have also to determine the values of the other three parameters. Let us consider the dynamic slider for instance. In the Figure 3, we have four images obtained from filtering in low level, with the default values of scale and scale division, but having different values of dynamic slider. In this case, the choice of the best image is not so evident. If we consider their entropies, we have the results given in the Figure 4. In this figure, it is also plotted the entropy of the original image as reference. Note that, in this case, two of the filtered image are worse than the original. However, the image obtained from low level and default values filtering has the highest entropy.

In the Figure 5, we considered also another case, that of different levels with a different value of scale, 16 instead of default value 240. As in the case of Fig.3, to have the best image we use the entropy as given in the plots of Figure 6. The entropy is maximized in the case of the low-level, scale 16 image. All these examples are showing that the entropy is sensitive in distinguishing the role of parameters. However, these examples are also telling that, when we decide to adjust all the parameters, it is necessary a calculation of the entropy on a bulk set of images, as we did in [7]. To this task, it will be devoted our future work.

**Appendix**

In the following formulas, we can see how the entropies of Shannon, Tsallis and Kaniadakis are defined. To apply these formulas to images, we have to consider $p_i$ as the frequency of grey tones. The index I is giving the specific grey tone. We have:

$$(A1) \quad \text{Shannon:} \quad S = -\sum_i p_i \ln p_i$$

$$(A2) \quad \text{Tsallis:} \quad T = T_q = \frac{1}{q-1}\left(1 - \sum_i p_i^q\right)$$

$$(A3) \quad \text{Kaniadakis} (\kappa\text{-entropy}): \quad K_\kappa = -\sum_i \frac{p_i^{1+\kappa} - p_i^{1-\kappa}}{2\kappa}$$

In (2A),(A3) we have entropic indices $q$ and $\kappa$. Note that $\lim_{q \to 1} T = S$; $\lim_{\kappa \to 0} K = S$.

If we have two independent subsystems A,B, the joint entropies H(A,B) are given by:

$$(A4) \quad S(A,B) = S(A) + S(B)$$

$$(A5) \quad T(A,B) = T(A) + T(B) + (1-q)T(A)T(B)$$

$$(A6) \quad K(A,B) = K(A)\Im(B) + K(B)\Im(A) \quad \text{with} \quad \Im = \sum_i \frac{p_i^{1+\kappa} + p_i^{1-\kappa}}{2}$$

Note that for the generalized additivity of $\kappa$-entropy, we need another function containing probabilities (see [10] and references therein). When we have a small value of the entropic index, function $\Im$ is equal to 1. For relations existing between Kaniadakis and Tsallis entropies, see please Ref.11.

The conditional entropies are given as follows. The conditional Tsallis entropy is [12]:

$$(A7) \quad T_q(A|B) = \frac{T_q(A,B) - T_q(B)}{1 + (1-q)T_q(B)}$$

Then $T_q(A|B)[1+(1-q)T_q(B)] = T_q(A,B) - T_q(B)$. The expression of Kaniadakis conditional entropy is [11]:

$$(A8) \quad K_\kappa(A|B) = \frac{K_\kappa(A,B) - K_\kappa(B)\Im_\kappa(A|B)}{\Im_\kappa(B)}$$

$$(A9) \quad \Im_\kappa(A|B) = \frac{\Im_\kappa(A,B) - (1-q)^2 K_\kappa(A|B) K_\kappa(B)}{\Im_\kappa(B)}$$

If we are using the entropic index with a low value, expression (A8) becomes:

$$(A10) \quad K_\kappa(A|B) = \frac{K_\kappa(A,B) - K_\kappa(B)\Im_\kappa(A)}{\Im_\kappa(B)}$$

The mutual entropy (without renormalization) is [11]:

$$(A11) \quad MK_\kappa(A;B) = K_\kappa(A) - K_\kappa(A|B) = \frac{K_\kappa(A)\Im_\kappa(B) - K_\kappa(A,B) + K_\kappa(B)\Im_\kappa(A|B)}{\Im_\kappa(B)}$$

When A and B are independent, $\Im(A|B) = \Im(A)$ and then the mutual information is zero.

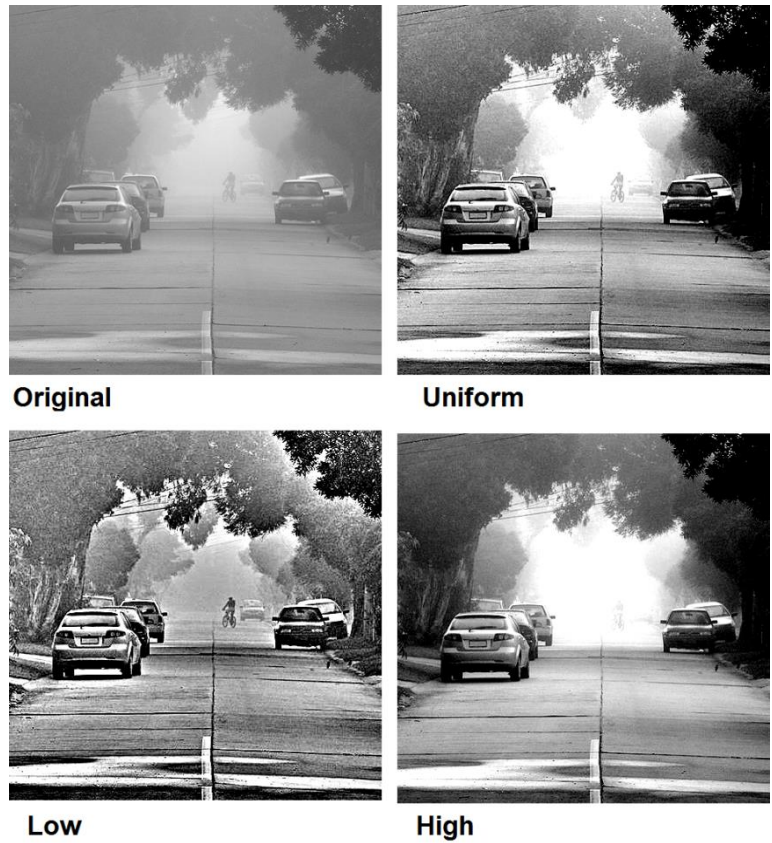

Figure 1: The original image is given in the left-upper panel, (Courtesy: Ian W. Fleggen, Wikipedia, 20880313, Foggy Street). The other images had been obtained using GIMP Retinex at uniform, low and high levels. The best level seems to be the low one.

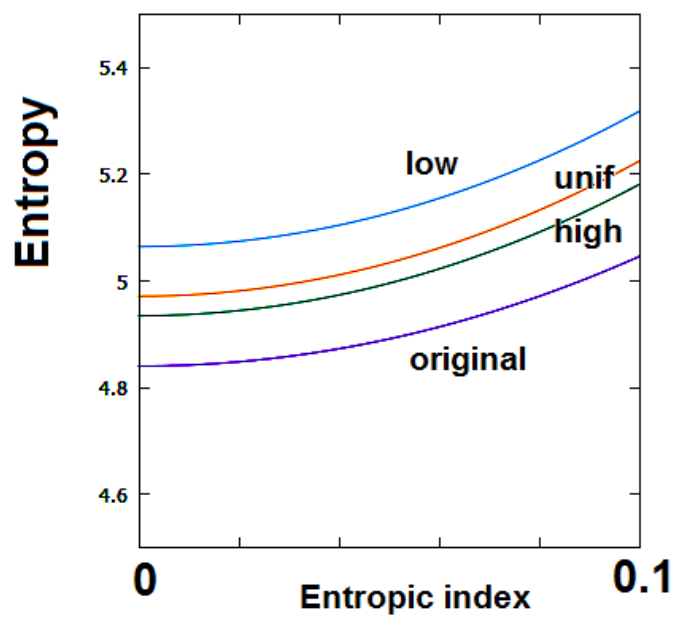

Figure 2: Of the images given in Fig.1, we can calculate the entropies. The plots show the Kaniadakis entropy as a function of its entropic index. When this index approaches zero, the Kaniadakis entropy is the Shannon entropy. The entropy is maximized in the case of the low-level filtered image.

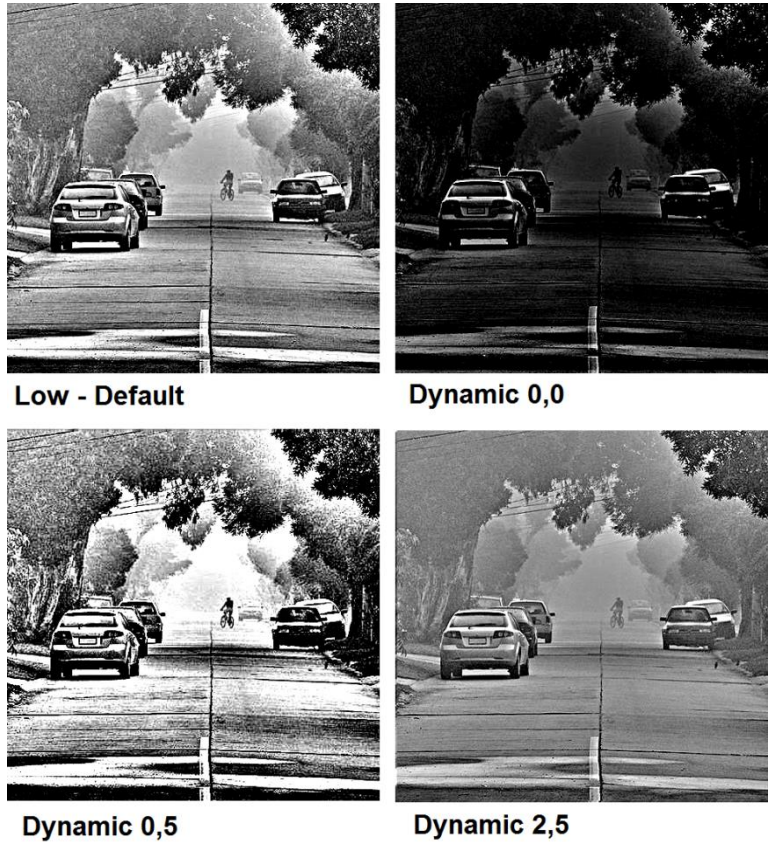

Figure 3: Low-level filtered images with four different values of dynamic slider. The default value is 1,2.

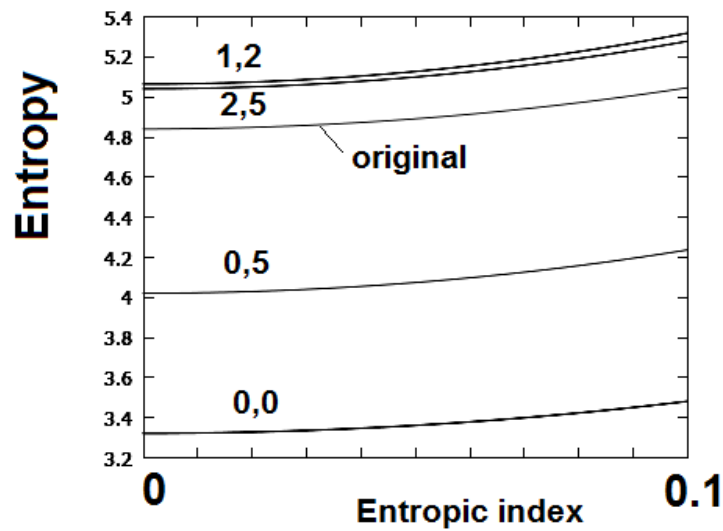

Figure 4: Of the images given in Fig.3, we can calculate the entropies. The plots show the Kaniadakis entropy as a function of its entropic index. When this index approaches zero, the Kaniadakis entropy is the Shannon entropy. The entropy is maximized in the case of the default value (1,2) of the dynamic slider. The entropy of the original image is also given for comparison.

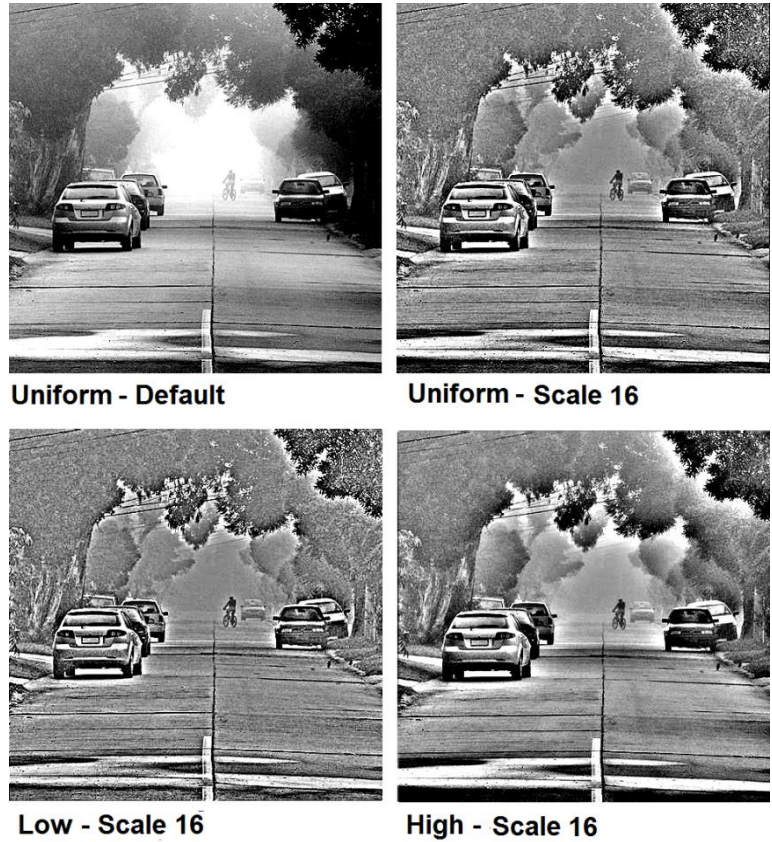

Figure 5: The different levels with the same value (16) of the "scale", but different from the default value (240). As in the case of Fig.3, to have the best image we use the entropy as given in the following plots.

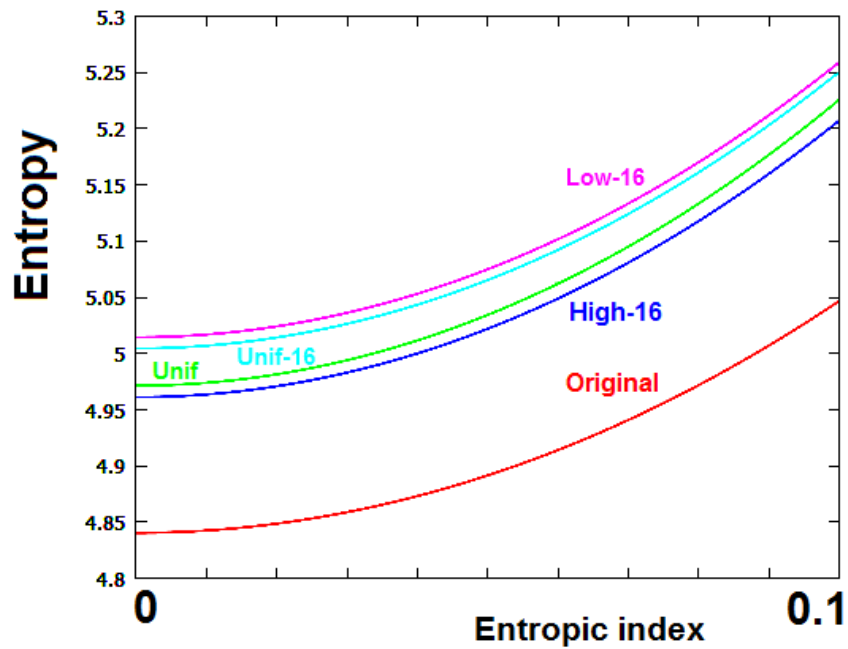

Figure 6: Of the images given in Fig.5, we can calculate the entropy. The plots show the Kaniadakis entropy as a function of its entropic index. When this index approaches zero, the Kaniadakis entropy is the Shannon entropy. The entropy is maximized in the case of the low-level, scale 16 image. The entropy of the original image is also given for comparison.